\documentclass[11pt,a4paper]{article}
\usepackage[hyperref]{emnlp-ijcnlp-2019}
\usepackage{times}
\usepackage{latexsym}

\usepackage{url}

\usepackage{booktabs}
\usepackage{graphicx}
\usepackage{amsmath}
\usepackage{amssymb}
\usepackage{multirow}
\usepackage{verbatim}
\usepackage{color}
\usepackage{empheq}

\aclfinalcopy 

\title{Question-type Driven Question Generation}

\author{Wenjie Zhou, Minghua Zhang, Yunfang Wu\thanks{Corresponding author.} \\
Key Laboratory of Computational Linguistics, Ministry of Education \\
School of Electronics Engineering and Computer Science, Peking University, Beijing, China \\
{\tt \{wjzhou013, zhangmh, wuyf\}@pku.edu.cn}
}

\date{}

\begin{document}
\maketitle
\begin{abstract}
Question generation is a challenging task which aims to ask a question based on an answer and relevant context. The existing works suffer from the mismatching between question type and answer, i.e. generating a question with type $how$ while the answer is a personal name. We propose to automatically predict the question type based on the input answer and context. Then, the question type is fused into a seq2seq model to guide the question generation, so as to deal with the mismatching problem. We achieve significant improvement on the accuracy of question type prediction and finally obtain state-of-the-art results for question generation on both SQuAD and MARCO datasets.
  
\end{abstract}

\section{Introduction}
Question generation (QG) can be effectively applied to many fields, including question answering \cite{DBLP:conf/emnlp/DuanTCZ17}, dialogue system \cite{DBLP:journals/jzusc/ShumHL18} and education. In this paper, we focus on the answer-aware QG, which is to generate a question according to the given sentence and the expected answer. 
	
Recently, the neural-based approaches on QG have achieved remarkable success, by applying large-scale reading comprehension datasets and employing the encoder-decoder framework. Most of the existing works are based on the seq2seq network incorporating attention mechanism and copy mode, which are first applied in \citet{DBLP:conf/nlpcc/ZhouYWTBZ17}. Later, \citet{DBLP:conf/naacl/SongWHZG18} leverage multi-perspective matching methods, and \citet{DBLP:conf/emnlp/SunLLHMW18} propose a position-aware model to put more emphasis on answer-surrounded context words. Both works are trying to enhance the relevance between the question and answer. \citet{DBLP:conf/emnlp/ZhaoNDK18} aggregate paragraph-level context to provide sufficient information for question generation. Another direction is to integrate question answering and question generation as dual tasks \cite{DBLP:journals/corr/TangDQZ17}. Beyond answer-aware QG, some systems try to generate questions from a text without answer as input \cite{DBLP:conf/emnlp/DuC17,DBLP:conf/acl/SubramanianWYZT18}. 

Despite the progress achieved by the previous work, we found that the types of generated questions are often incorrect. According to experiments on SQuAD, one strong model \cite{DBLP:conf/nlpcc/ZhouYWTBZ17} we replicate only obtains 57.6\% accuracy in question type. As we know, question type is vital for question generation, since it determines the question pattern and guides the generating process. If the question type is incorrect, the remained generated sequence would drift far away.

\begin{figure*}
\centering
\includegraphics[scale=0.525]{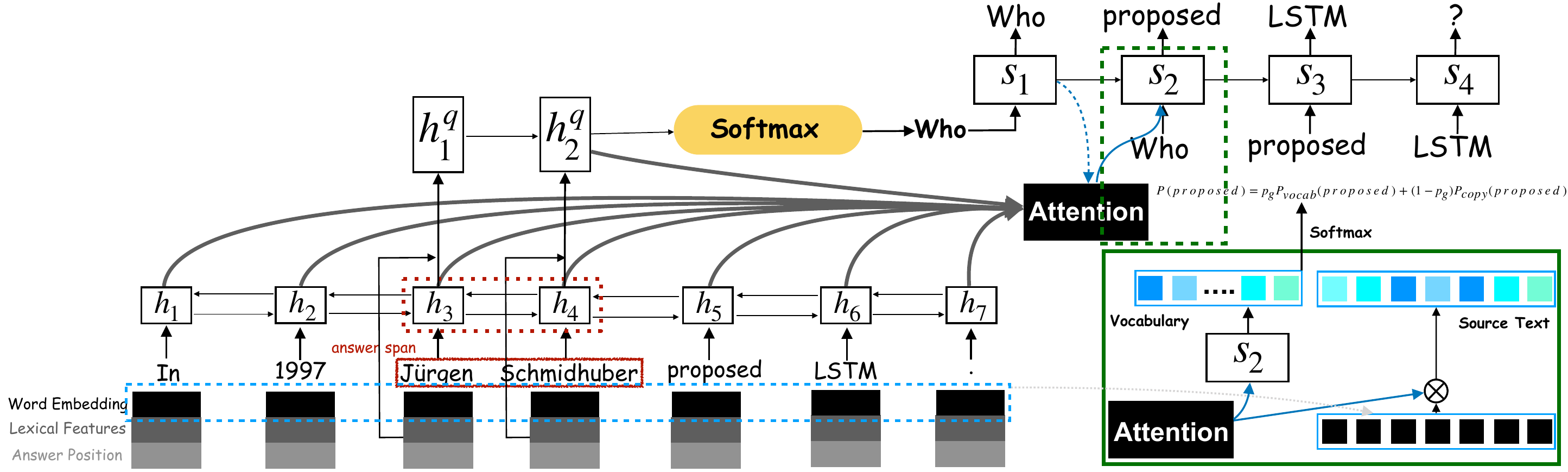}
\caption{\label{unified model} Structure of our unified model}
\end{figure*}

Several works have addressed this issue. \citet{DBLP:conf/emnlp/SunLLHMW18} incorporated a question word generation mode to generate question word at each decoding step, which utilized the answer information by employing the encoder hidden states at the answer start position. However, their method did not consider the structure and lexical features of answer. Meanwhile, the way they utilize the question word is not as effective as ours. \citet{DBLP:conf/iclr/HuLM0Y18} proposed a model to generate question based on the given question type and aspect, and their work verifies the effect of question word but fails in the conventional QG task which does not give a question type, since their experimental results show poor performance when trying to generate question types automatically. Our work solves this problem as Section \ref{my section} shows. \citet{DBLP:conf/acl/HuangNWL18} devised a type decoder which combines three type-specific generation distribution (including question type) with weighted sum. However, the results displayed in their paper show that questions in dialogue are far different from questions for reading comprehension, which indicates a gap between two tasks.

In this paper, we propose a unified model to predict the question type and to generate questions simultaneously. We conduct experiments on two reading comprehension datasets: SQuAD and MARCO, and obtain promising results. As for the auxiliary task, our unified model boosts the accuracy of question type prediction significantly, by 16.79\% on SQuAD and 3.5\% on MARCO. For question generation, our model achieves new state-of-the-art results on both datasets, with BLEU-4 16.31 on SQuAD and 21.59 on MARCO.     

\section{Model Description}
The structure of our model is shown in Figure \ref{unified model}. A feature-rich encoder is used to encode the input sentence and corresponding answer that is a span of the sentence. Besides, the answer hidden states are used to predict the type of target question. This prediction will further be used to guide QG with a unified attention-based decoder.

\subsection {Feature-rich Encoder}
Follow \citet{DBLP:conf/nlpcc/ZhouYWTBZ17}, we exploit lexical features to enrich the encoder, where features are composed of POS tags, NER tags, and word case. We concatenate the word embedding $e_t$, answer position embedding $a_t$ and lexical features embedding $l_t$ as input ($(x_1, ...,x_T)$, $x_t = [e_t;a_t;l_t]$). Then a bidirectional LSTM is used to produce a sequence of hidden states $(h_1, ..., h_T)$.

\subsection{Question Type Prediction}
Since different types of questions are various in syntax and semantics, it is essential to predict an accurate type for generating a reasonable question. According to the statistics on SQuAD, 78\% of questions in the training set begin with the 7 most common used question words as Table \ref{tab:type} shows. So we divide question types into 8 categories, including 7 question words and an additional type 'others'. It shows that the data distribution on different types is quite unbalanced, suggesting it is a hard task to predict the correct question type.     

\begin{table*}[!ht]
\centering
\small
\resizebox{0.75\textwidth}{!}{
\begin{tabular}{l@{\
}|c@{\quad}c@{\quad}c@{\quad}c@{\quad}c@{\quad}c@{\quad}c@{\quad}c@{\quad}}
\toprule
\textbf{Type} & \textbf{What} & \textbf{Who} & \textbf{How} & \textbf{When} & \textbf{Which} & \textbf{Where} & \textbf{Why} & \textbf{Others} \\
\midrule
\textbf{SQuAD} & 43.26\% & 9.39\% & 9.12\% & 6.26\% & 4.78\% & 3.76\% & 1.37\% & 21.83\% \\
\textbf{MARCO} & 44.29\% & 1.47\% & 16.28\% & 1.52\% & 1.11\% & 4.16\% & 1.88\% & 29.29\% \\
\bottomrule
\end{tabular}}
\caption{Proportions of each type of questions on two datasets.}
\label{tab:type}
\end{table*}

We use an unidirectional LSTM network to predict the expected question type. Assuming \textit{m+1, ..., m+a} is the index of given answer span in the input sentence, based on the corresponding feature-rich hidden states $(h_{m+1}, ..., h_{m+a})$, we calculate the question type hidden states as follow:
\begin{align}
\label{q lstm} 
h_j^q\!=\!LSTM^q & ([h_{m\!+\!j}; l_{m\!+\!j}], h_{j\!-\!1}^q), j\!\in\![1, a] \\
& h_0^q = h_T
\end{align}
where $l_{m+j}$ is the corresponding lexical features. Besides, to make full use of the feature-rich encoder hidden states, we take the last hidden state as the initial hidden state of $LSTM^q$.

Then, the last output hidden state $h_a^q$ are fed into a softmax layer to obtain the type distribution:
\begin{align}
\label{q softmax}
P(Q_w) & = softmax(W_q h_a^q) \\
E^{q} & = - log(P(Q_w^*))
\end{align}
$E^q$ is the loss of question type prediction, $Q_w^*$ is the target type.

\subsection{Unified Attention-based Decoder}
The conventional attention-based decoder adopts a $<$BOS$>$ token as the first input word at step $1$, while we replace it with the predicted question word $Q_w$ to guide the generation process. At step $i$, we calculate the decoder hidden state as follow:
\begin{align}
s_{i} & = LSTM([w_{i-1};c_{i-1}], s_{i-1})
\end{align}

Further, to fuse the type information in every step of question generation, our model takes $h_a^q$ into consideration while calculating the context vector, i.e. the context vector is conditioned on $(h_1, ..., h_T, h_{T+1})$, where $h_{T+1} = h_a^q$.
\begin{align}
\label{context vector}
c_i & = \sum\limits_{t=1}^{T+1} {\alpha}_{it}h_t
\end{align}
where ${\alpha}_{it}$ is calculated via attention mechanism \cite{DBLP:journals/corr/BahdanauCB14}.

Then, $s_i$ together with $c_i$ will be fed into a two-layer feed-forward network to produce the vocabulary distribution $P_{vocab}$.

Following \cite{DBLP:conf/acl/SeeLM17}, the copy mode is used to duplicate words from the source via pointing:
\begin{align}
P(w_i) & \!=\! p_{g}P_{vocab}(w_i) \!+\! (1\!-\!p_{g})P_{copy}(w_i)
\end{align}
where $P_{copy}$ is the distribution of copy mode, $p_{g}\in{[0, 1]}$ is a gate to dynamically assign weights.

The loss at step $i$ is the negative log-likelihood of the target word $w_i^*$. To obtain a combined training objective of two tasks, we add the loss of question type prediction into the loss of question generation to form a total loss function:

\begin{align}
E^{total} = \frac{1}{K} \sum\limits_{i=1}^{K} -logP(w_i^*) + E^{q}
\end{align}

\section{Experiment}

\subsection{Experiment Settings}
\textbf{Dataset} Following the previous works, we conduct experiments on two datasets, SQuAD and MARCO. We use the data released by \citet{DBLP:conf/nlpcc/ZhouYWTBZ17} and \citet{DBLP:conf/emnlp/SunLLHMW18}, there are 86,635, 8,965 and 8,964 sentence-answer-question triples in the training, development and test set for SQuAD, and 74,097, 4,539 and 4,539 sentence-answer-question triples in the training, development and test set for MARCO, respectively. Lexical features are extracted using Stanford CoreNLP.

\noindent \textbf{Implementation Details} Our vocabulary is set to contain the most frequent 20,000 words in each training set. Word embeddings are initialized with the pre-trained 300-dimensional Glove vectors, and they are allowed to be fine-tuned during training. The representations of answer position, POS tags, NER tags, and word case are randomly initialized as 32-dimensional vectors, respectively. The feature-rich encoder consists of 2 layers BiLSTM, and the hidden size of all encoders and decoder is set to 512. The cutoff length of the input sequences is set to 10. In testing, we used beam search with a beam size of 12. The development set is used to search the best checkpoint. In order to decrease the volatility of the training procedure, we then average the nearest 5 checkpoints to obtain a single averaged model. 

Following the existing work, we use BLUE \cite{DBLP:conf/acl/PapineniRWZ02} as the metrics for automatic evaluation, with BLUE-4 as the main metric.

\subsection{Upper Bound Analysis}
To study the effectiveness of the question type for QG, we make an upper bound analysis. The experimental results are shown in Table \ref{tab:upper bound}. 

\begin{table*}[!ht]
\centering
\resizebox{\textwidth}{!}{
\begin{tabular}{l@{\
}|c@{\quad}c@{\quad}c@{\quad}c@{\quad}c@{\quad}c@{\quad}c@{\quad}c@{\quad}}
\toprule
\textbf{Dataset} & \multicolumn{4}{c}{\textbf{SQuAD}} & \multicolumn{4}{c}{\textbf{MARCO}} \\
\textbf{Model} & \textbf{BLEU-1} & \textbf{BLEU-2} & \textbf{BLEU-3} & \textbf{BLEU-4} & \textbf{BLEU-1} & \textbf{BLEU-2} & \textbf{BLEU-3} & \textbf{BLEU-4} \\
\midrule
\midrule
feature-rich pointer generator (baseline) & 41.25 & 26.76 & 19.53 & 14.89 & 54.04 & 36.68 & 26.62 & 20.15 \\
\midrule
given the first word of question & 47.00 & 31.82 & 23.69 & 18.30  & 59.51 & 41.42 & 30.70 & 23.47 \\
given the question type & 45.56 & 30.53 & 22.61 & 17.38  & 57.39 & 39.60 & 29.29 & 22.52 \\ 
\bottomrule
\end{tabular}}
\caption{Upper bound analysis by incorporating question words with different ways.}
\label{tab:upper bound}
\end{table*}

\begin{table*}[!ht]
\centering
\resizebox{\textwidth}{!}{
\begin{tabular}{l@{\
}|c@{\quad}c@{\quad}c@{\quad}c@{\quad}c@{\quad}c@{\quad}c@{\quad}c@{\quad}}
\toprule
\textbf{Dataset} & \multicolumn{4}{c}{\textbf{SQuAD}} & \multicolumn{4}{c}{\textbf{MARCO}} \\
\textbf{Model} & \textbf{BLEU-1} & \textbf{BLEU-2} & \textbf{BLEU-3} & \textbf{BLEU-4} & \textbf{BLEU-1} & \textbf{BLEU-2} & \textbf{BLEU-3} & \textbf{BLEU-4} \\
\midrule
\midrule
NQG++ \cite{DBLP:conf/nlpcc/ZhouYWTBZ17} & - & - & - & 13.29 & - & - & - & - \\
question word generate model \cite{DBLP:conf/emnlp/SunLLHMW18} & \ 42.10 & 27.52 & 20.14 & 15.36 & 46.59 & 33.46 & 24.57 & 18.73 \\
Hybrid model \cite{DBLP:conf/emnlp/SunLLHMW18} & {43.02} & {28.14} & {20.51} & {15.64} & 48.24 & 35.95 & 25.79 & 19.45 \\
Maxout Pointer (sentence) \cite{DBLP:conf/emnlp/ZhaoNDK18} & \textbf{44.51} & 29.07 & {21.06} & 15.82 & - & - & - & 16.02 \\
\midrule
\midrule
\multicolumn{9}{l}{\textbf{Our Model}} \\
\midrule
Feature-rich pointer generator (Baseline) & 41.25 & 26.76 & 19.53 & 14.89 & 54.04 & 36.68 & 26.62 & 20.15 \\
Question-type driven model (Unified-model) & 43.11 & \textbf{29.13} & \textbf{21.39} & \textbf{16.31}  & \textbf{55.67} & \textbf{38.16} & \textbf{28.12} & \textbf{21.59} \\
\bottomrule
\end{tabular}}
\caption{Experimental results of our model comparing with previous methods on two datasets.}
\label{tab:results}
\end{table*}

First, we feed the decoder with the original first word of questions, regardless whether the first word is a question word or not. As shown in Table \ref{tab:upper bound}, comparing with the baseline model, the performance gets 3.41 and 3.32 points increment on SQuAD and MARCO, respectively, which is a large margin. 

Since the beginning words which are not question words compose a large vocabulary while the number of each word is few, it is irrational to train a classifier to predict all types of these beginning words. 
Therefore, we reduce the question type vocabulary to only question words. In details, if the start of a targeted question is a question word, the corresponding question word will replace the $<$BOS$>$ to feed into decoder, otherwise we just use the original $<$BOS$>$ without any replacement. This experiment still gains a lot, as shown in Table \ref{tab:upper bound} with "given the quetion type".

The above experiments verify the magnitude of using the proper question type to guide the generation process, suggesting us a promising direction for QG. 

\subsection{Results and Analysis}
\label{my section}
\textbf{Main Results} The experimental results on two datasets are shown in Table \ref{tab:results}. By incorporating the question type prediction, our model obtains obvious performance gain over the baseline model, with 1.42 points gain on SQuAD and 1.44 on MARCO. Comparing with previous methods, our model outperforms the existing best methods, achieving new state-of-the-art results on both datasets, with 16.31 on SQuAD and 21.59 on MARCO. 

\noindent \textbf{Question Type Accuracy} We evaluate different models in terms of Beginning Question Word Accuracy (BQWA). This metric measures the ratio of the generated questions that share the same beginning word with the references which begin with a question word. Table \ref{tab:BQWA} displays the BQWA of two models on both datasets. It shows that our unified model brings significant performance gain in question type prediction.   

\begin{figure*}
\centering
\includegraphics[scale=0.35]{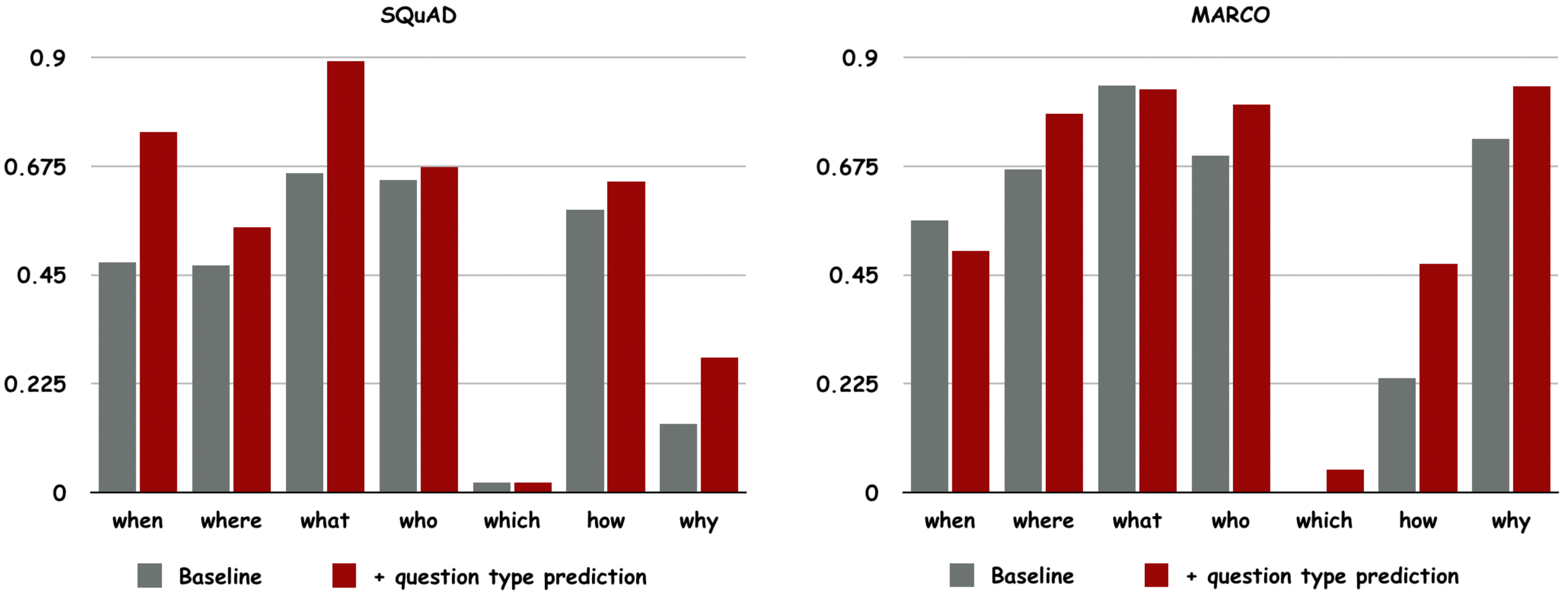}
\caption{\label{type_rate} The accuracy on different question words.}
\end{figure*}

Further, in Figure \ref{type_rate} we show the accuracy of different question words on both datasets in detail. Our unified model outperforms the baseline for all question types on SQuAD, and all but two types on MARCO.

\begin{table}[!ht]
\centering
\small
\resizebox{.48\textwidth}{!}{
\begin{tabular}{l@{\
}|c@{\quad}c@{\quad}}
\toprule
\textbf{Model} & \textbf{BQWA on SQuAD} & \textbf{BQWA on MARCO} \\
\midrule
Baseline & 57.62\% & 76.98\% \\
Unified model & \textbf{74.41\%} & \textbf{80.48\%} \\
\bottomrule
\end{tabular}}
\caption{Experiments of the beginning question word accuracy.}
\label{tab:BQWA}
\end{table}

\noindent \textbf{Model Analysis} We conduct experiments on different variants of our model, as shown in Table \ref{tab:ablation study}. ``w/o answer hidden state'' takes $(x_{m+1}, ..., x_{m+a})$ as the input of type decoder instead of answer hidden states; ``w/o question word replace $<$BOS$>$'' simply use $<$BOS$>$ as the first input word of decoder. Experiments on SQuAD verify the effectiveness of our model setting. 

\begin{table}[!ht]
\centering
\resizebox{.48\textwidth}{!}{
\begin{tabular}{l@{\
}|c@{\quad}c@{\quad}c@{\quad}c@{\quad}}
\toprule
\textbf{Dataset} & \multicolumn{4}{c}{\textbf{SQuAD}} \\
\textbf{Model} & \textbf{BLEU-1} & \textbf{BLEU-2} & \textbf{BLEU-3} & \textbf{BLEU-4} \\
\midrule
\midrule
Unified-model & 43.11 & 29.13 & 21.39 & 16.31 \\
w/o answer hidden states & 41.91 & 27.77 & 20.37 & 15.54 \\
w/o question word replace $<$BOS$>$ & 41.05 & 26.93 & 19.73 & 15.10 \\
\bottomrule
\end{tabular}}
\caption{Experiments on different settings of our model.}
\label{tab:ablation study}
\end{table}

\noindent \textbf{Case Study} To show the effect of question words prediction on question generation, Table \ref{tab:case} lists some typical examples.

In the first example, the baseline fails to recognize \emph{Len-shaped} as an adjective, while the unified-model succeeds by utilizing lexical features which are the input of question type prediction layer. 

In the second example, the baseline assigns a location type for the generated question based on \emph{American} and \emph{Israel}, it fails to consider the whole answer span. Our unified model resolves it by encoding the answer hidden states sequence as a whole. 

The third example shows another typical error, which fails to consider answer surrounding context. In this example the given answer only contains a number \emph{66}. By taking \emph{article 66} into account, we know the question should not be a numerical type. Since the answer hidden states used as the input of type prediction contain surrounding information, our model resolves this problem.

\begin{table}[!ht]\small
    \centering
    \begin{tabular}{p{7.25cm}}
    \toprule
        \textbf{Context:} In land plants , chloroplasts are generally \underline{lens-shaped}, 5--8 μm in diameter and 1--3 μm thick. \\
        \textbf{Reference:} How are chloroplasts in land plants usually shaped? \\
        \textbf{Baseline:} What are chloroplasts generally generally? \\
        \textbf{Unified-model:} How are chloroplasts in land plants? \\
        \midrule
        \textbf{Context:} \underline{In response to American aid to Israel}, on October 16, 1973, OPEC raised the posted price of oil by 70\%, to \$5.11 a barrel. \\
        \textbf{Reference:} Why did the oil ministers agree to a cut in oil production? \\
        \textbf{Baseline:} Where did OPEC receive the price of oil by 70\%? \\
        \textbf{Unified-model:} Why did OPEC raised the posted price of oil by 70\%? \\
        \midrule
        \textbf{Context:} Article 65 of the agreement banned cartels and article \underline{66} made provisions for concentrations, or mergers, and the abuse of a dominant position by companies. \\
        \textbf{Reference:} Which article made provisions for concentrations or mergers and the abuse of a dominant position by companies? \\
        \textbf{Baseline:} How many provisions made provisions for concentrations? \\
        \textbf{Unified-model:} Which article made provisions for concentrations, or mergers? \\
    \bottomrule
    \end{tabular}
    \caption{Examples of generated questions. The underline words are the target answer.}
    \label{tab:case}
\end{table}

\section{Conclusion}
In this paper, we discuss the challenge in question type prediction for question generation. We propose a unified model to predict the type and utilize it to guide the generation of question. Experiments on SQuAD and MARCO datasets verify the effectiveness of our model. We improve the accuracy of question type prediction by a large margin, and achieve new state-of-the-art results for question generation.  

\section*{Acknowledgments}
We thank Weikang Li, Xin Jia and Nan Jiang for their valuable comments and suggestions. This work is supported by the National Natural Science Foundation of China (61773026).

\bibliography{emnlp-ijcnlp-2019}

\begin{thebibliography}{14}
\expandafter\ifx\csname natexlab\endcsname\relax\def\natexlab#1{#1}\fi

\bibitem[{Bahdanau et~al.(2014)Bahdanau, Cho, and
  Bengio}]{DBLP:journals/corr/BahdanauCB14}
Dzmitry Bahdanau, Kyunghyun Cho, and Yoshua Bengio. 2014.
\newblock \href {http://arxiv.org/abs/1409.0473} {Neural machine translation by
  jointly learning to align and translate}.
\newblock \emph{CoRR}, abs/1409.0473.

\bibitem[{Du and Cardie(2017)}]{DBLP:conf/emnlp/DuC17}
Xinya Du and Claire Cardie. 2017.
\newblock \href {https://aclanthology.info/papers/D17-1219/d17-1219}
  {Identifying where to focus in reading comprehension for neural question
  generation}.
\newblock In \emph{Proceedings of the 2017 Conference on Empirical Methods in
  Natural Language Processing, {EMNLP} 2017, Copenhagen, Denmark, September
  9-11, 2017}, pages 2067--2073.

\bibitem[{Duan et~al.(2017)Duan, Tang, Chen, and
  Zhou}]{DBLP:conf/emnlp/DuanTCZ17}
Nan Duan, Duyu Tang, Peng Chen, and Ming Zhou. 2017.
\newblock \href {https://aclanthology.info/papers/D17-1090/d17-1090} {Question
  generation for question answering}.
\newblock In \emph{Proceedings of the 2017 Conference on Empirical Methods in
  Natural Language Processing, {EMNLP} 2017, Copenhagen, Denmark, September
  9-11, 2017}, pages 866--874.

\bibitem[{Hu et~al.(2018)Hu, Liu, Ma, Zhao, and Yan}]{DBLP:conf/iclr/HuLM0Y18}
Wenpeng Hu, Bing Liu, Jinwen Ma, Dongyan Zhao, and Rui Yan. 2018.
\newblock \href {https://openreview.net/forum?id=rkRR1ynIf} {Aspect-based
  question generation}.
\newblock In \emph{6th International Conference on Learning Representations,
  {ICLR} 2018, Vancouver, BC, Canada, April 30 - May 3, 2018, Workshop Track
  Proceedings}.

\bibitem[{Papineni et~al.(2002)Papineni, Roukos, Ward, and
  Zhu}]{DBLP:conf/acl/PapineniRWZ02}
Kishore Papineni, Salim Roukos, Todd Ward, and Wei{-}Jing Zhu. 2002.
\newblock \href {http://www.aclweb.org/anthology/P02-1040.pdf} {Bleu: a method
  for automatic evaluation of machine translation}.
\newblock In \emph{Proceedings of the 40th Annual Meeting of the Association
  for Computational Linguistics, July 6-12, 2002, Philadelphia, PA, {USA.}},
  pages 311--318.

\bibitem[{See et~al.(2017)See, Liu, and Manning}]{DBLP:conf/acl/SeeLM17}
Abigail See, Peter~J. Liu, and Christopher~D. Manning. 2017.
\newblock \href {https://doi.org/10.18653/v1/P17-1099} {Get to the point:
  Summarization with pointer-generator networks}.
\newblock In \emph{Proceedings of the 55th Annual Meeting of the Association
  for Computational Linguistics, {ACL} 2017, Vancouver, Canada, July 30 -
  August 4, Volume 1: Long Papers}, pages 1073--1083.

\bibitem[{Shum et~al.(2018)Shum, He, and Li}]{DBLP:journals/jzusc/ShumHL18}
Heung{-}Yeung Shum, Xiaodong He, and Di~Li. 2018.
\newblock \href {https://doi.org/10.1631/FITEE.1700826} {From eliza to xiaoice:
  challenges and opportunities with social chatbots}.
\newblock \emph{Frontiers of {IT} {\&} {EE}}, 19(1):10--26.

\bibitem[{Song et~al.(2018)Song, Wang, Hamza, Zhang, and
  Gildea}]{DBLP:conf/naacl/SongWHZG18}
Linfeng Song, Zhiguo Wang, Wael Hamza, Yue Zhang, and Daniel Gildea. 2018.
\newblock \href {https://aclanthology.info/papers/N18-2090/n18-2090}
  {Leveraging context information for natural question generation}.
\newblock In \emph{Proceedings of the 2018 Conference of the North American
  Chapter of the Association for Computational Linguistics: Human Language
  Technologies, NAACL-HLT, New Orleans, Louisiana, USA, June 1-6, 2018, Volume
  2 (Short Papers)}, pages 569--574.

\bibitem[{Subramanian et~al.(2018)Subramanian, Wang, Yuan, Zhang, Trischler,
  and Bengio}]{DBLP:conf/acl/SubramanianWYZT18}
Sandeep Subramanian, Tong Wang, Xingdi Yuan, Saizheng Zhang, Adam Trischler,
  and Yoshua Bengio. 2018.
\newblock \href {https://aclanthology.info/papers/W18-2609/w18-2609} {Neural
  models for key phrase extraction and question generation}.
\newblock In \emph{Proceedings of the Workshop on Machine Reading for Question
  Answering@ACL 2018, Melbourne, Australia, July 19, 2018}, pages 78--88.

\bibitem[{Sun et~al.(2018)Sun, Liu, Lyu, He, Ma, and
  Wang}]{DBLP:conf/emnlp/SunLLHMW18}
Xingwu Sun, Jing Liu, Yajuan Lyu, Wei He, Yanjun Ma, and Shi Wang. 2018.
\newblock \href {https://aclanthology.info/papers/D18-1427/d18-1427}
  {Answer-focused and position-aware neural question generation}.
\newblock In \emph{Proceedings of the 2018 Conference on Empirical Methods in
  Natural Language Processing, Brussels, Belgium, October 31 - November 4,
  2018}, pages 3930--3939.

\bibitem[{Tang et~al.(2017)Tang, Duan, Qin, and
  Zhou}]{DBLP:journals/corr/TangDQZ17}
Duyu Tang, Nan Duan, Tao Qin, and Ming Zhou. 2017.
\newblock \href {http://arxiv.org/abs/1706.02027} {Question answering and
  question generation as dual tasks}.
\newblock \emph{CoRR}, abs/1706.02027.

\bibitem[{Wang et~al.(2018)Wang, Liu, Huang, and
  Nie}]{DBLP:conf/acl/HuangNWL18}
Yansen Wang, Chenyi Liu, Minlie Huang, and Liqiang Nie. 2018.
\newblock \href {https://aclanthology.info/papers/P18-1204/p18-1204} {Learning
  to ask questions in open-domain conversational systems with typed decoders}.
\newblock In \emph{Proceedings of the 56th Annual Meeting of the Association
  for Computational Linguistics, {ACL} 2018, Melbourne, Australia, July 15-20,
  2018, Volume 1: Long Papers}, pages 2193--2203.

\bibitem[{Zhao et~al.(2018)Zhao, Ni, Ding, and Ke}]{DBLP:conf/emnlp/ZhaoNDK18}
Yao Zhao, Xiaochuan Ni, Yuanyuan Ding, and Qifa Ke. 2018.
\newblock \href {https://aclanthology.info/papers/D18-1424/d18-1424}
  {Paragraph-level neural question generation with maxout pointer and gated
  self-attention networks}.
\newblock In \emph{Proceedings of the 2018 Conference on Empirical Methods in
  Natural Language Processing, Brussels, Belgium, October 31 - November 4,
  2018}, pages 3901--3910.

\bibitem[{Zhou et~al.(2017)Zhou, Yang, Wei, Tan, Bao, and
  Zhou}]{DBLP:conf/nlpcc/ZhouYWTBZ17}
Qingyu Zhou, Nan Yang, Furu Wei, Chuanqi Tan, Hangbo Bao, and Ming Zhou. 2017.
\newblock \href {https://doi.org/10.1007/978-3-319-73618-1\_56} {Neural
  question generation from text: {A} preliminary study}.
\newblock In \emph{Natural Language Processing and Chinese Computing - 6th
  {CCF} International Conference, {NLPCC} 2017, Dalian, China, November 8-12,
  2017, Proceedings}, pages 662--671.

\end{thebibliography}
\bibliographystyle{acl_natbib}

\appendix

\end{document}